\documentclass[cameraready]{Interspeech}
\usepackage[]{todonotes} %disable to disable
\makeatletter
\newcommand*\iftodonotes{\if@todonotes@disabled\expandafter\@secondoftwo\else\expandafter\@firstoftwo\fi}  % defines \iftodonotes{<true>}{<false>}, thanks to \makeatother

\everypar{\looseness=-1}

\usepackage{xcolor}
\usepackage{tcolorbox}
\usepackage{csquotes}
\usepackage[table]{xcolor}

\definecolor{chartburgundy}{RGB}{115, 35, 70} % The deep burgundy/plum of the bars
\definecolor{quotebg}{RGB}{249, 246, 248}     % A very subtle, crisp off-white for the background
\definecolor{surveyblue}{RGB}{24, 64, 107} % Deep, sophisticated blue
\definecolor{surveybg}{RGB}{247, 249, 250}

\newtcolorbox{participantquote}{
    colback=quotebg,
    colframe=chartburgundy,
    leftrule=3pt,          % A tasteful, medium-thickness left line
    toprule=0pt,           % No top border
    bottomrule=0pt,        % No bottom border
    rightrule=0pt,         % No right border
    arc=0pt,               % Sharp, academic corners
    boxsep=5pt,            % Internal padding
    left=8pt,
    right=8pt,
    top=5pt,
    bottom=5pt
}

\newtcolorbox{slpquote}{
    colback=surveybg,
    colframe=surveyblue,
    leftrule=3pt,          % A tasteful, medium-thickness left line
    toprule=0pt,           % No top border
    bottomrule=0pt,        % No bottom border
    rightrule=0pt,         % No right border
    arc=0pt,               % Sharp, academic corners
    boxsep=5pt,            % Internal padding
    left=8pt,
    right=8pt,
    top=5pt,
    bottom=5pt
}

\title{Aligning Stuttered-Speech Research with End-User Needs:\\ Scoping Review, Survey, and Guidelines}
\author[affiliation={1}, orcid=0009-0008-3927-526X]{Hawau Olamide}{Toyin}
\author[orcid=0009-0002-1581-1220]{Mutiah}{Apampa}
\author[affiliation={1}, orcid=0000-0002-0361-7466]{Toluwani}{Aremu}
\author[affiliation={1}]{Humaid}{Alblooshi} 

\author[affiliation={2,5}, orcid=0000-0002-0511-554X]{Ana Rita}{Valente}
\author[affiliation={2}, orcid=0009-0007-2351-4791]{Gonçalo}{Leal}

\author[affiliation={3}, orcid=0000-0002-1101-549X]{Zhengjun}{Yue}
\author[affiliation={4}, orcid=0000-0001-5503-867X]{Zeerak}{Talat}
\author[affiliation={1}, orcid=0000-0003-1706-1777]{Hanan}{Aldarmaki}

\address{
  $^1$MBZUAI, UAE; $^2$SpeechCare, Portugal \& UAE; $^3$SLAI \& CUHK (SZ), China; \\
  $^4$University of Edinburgh, UK; $^5$School of Health Sciences \& IEETA, University of Aveiro, Portugal }

\email{\{hawau.toyin,hanan.aldarmaki\}@mbzuai.ac.ae}


\usepackage{comment}

\usepackage{svg}
\usepackage{graphicx}
\usepackage{hyperref}
\usepackage{url}
\usepackage{dirtytalk}
\usepackage{csquotes}
\usepackage{booktabs}
\usepackage{multirow}
\usepackage{tabularx}
\usepackage{arydshln}

\usepackage[suppress]{color-edits} % [suppress]for getting rid of comments
\addauthor{zt}{olive}

\usepackage{xspace}

\newcommand{\dq}[1]{``#1''}
\newcommand{\slp}{\texttt{SLPs}\xspace}
\newcommand{\pws}{\texttt{PWS}\xspace}

\usepackage{footmisc}
\begin{document}

\maketitle

\begin{abstract}
   % Atypical speech research has received significant attention from the research community in recent years, however with limited avenues for inter-disciplinary discussion. While we know there are limitations in current state of research across different areas, it's unclear the extent of these limitations and what effects they have on stakeholder needs, if any. In this paper, we first address these questions through findings from a 228 stuttered-speech AI paper scoping review and find gaps relating to naming standards,task formulation, open-science and gaps in stakeholder needs. In a 70-person survey we identify priorities, pain-points and needs of stakeholders from voice-based AI.
   % In this work we seek to examine the needs of atypical speech stakeholders and identify alignment/misalignment with research practices. We subsequently align findings from the review and survey to find the alignment gap between stakeholders and current state of research. Our study provides actionable insights for truly user-centric stuttered-speech research.

Atypical speech is receiving greater attention in speech technology research, but much of this work unfolds with limited interdisciplinary dialogue. For stuttered speech in particular, it is widely recognised that current speech recognition systems fall short in practice, and current evaluation methods and research priorities are not systematically grounded in end-user experiences and needs. 
In this work, we analyse these gaps through 1) a scoping review of papers that deal with stuttered speech and 2) a survey of 70 stakeholders, including adults who stutter and speech-language pathologists. By analysing these two perspectives, we propose a taxonomy of stuttered-speech research, identify where current research directions diverge from the needs articulated by stakeholders, and conclude by outlining concrete guidelines and directions towards addressing the real needs of the stuttering community.

\end{abstract}
\section{Introduction}
% In recent years, there has been an increase in research activity at the intersection of speech technology and atypical speech, such as stuttered-speech analysis, e.g., in automatic speech recognition (ASR), speech synthesis, and event classification. 
In recent years, research at the intersection of speech technology and atypical speech, including stuttering, has increased, with applications of automatic speech recognition (ASR)~\cite{Mujtaba2024InclusiveAF}, speech synthesis~\cite{Bhat2023AdversarialTF}, and event classification  \cite{valente25_interspeech}.
Stuttering is a variable, multidimensional speech disorder that disrupts speech fluency, leading to the occurrence of different \emph{stuttering events}, which are disruptions in the flow of speech, which can typically be described as repetitions, prolongations or blocks~\cite{valente25_interspeech}. 
These events vary among People Who Stutter (henceforth, \pws) in intensity and frequency, and can be accompanied by other coping behaviours, such as facial grimaces. %and are typically used to measure stuttering severity and treatment progress tracking.
Stuttering events are thus inherently multi-modal, extending beyond verbal disfluencies to include facial and extremity movements. In clinical practice, Speech-Language Pathologists (henceforth, \slp) must quantify and characterize such events to detect, diagnose, assess, and develop treatment plans. In speech technology literature, however, the terms \emph{\dq{stuttering}} and \emph{\dq{disfluency}} are often used interchangeably to describe the stuttering events observable in the \emph{produced speech only}. 
Recent work by Valente et al.~\cite{valente25_interspeech} contributed new clinically informed multi-modal annotations for stuttering events, illustrating both the importance and the difficulty of agreeing on operational definitions for computational modelling of stuttered speech.
Despite growing interest in atypical speech processing, qualitative studies of everyday failures~\cite{Lea2023FromUP,Li2025OurCV} and quantitative evaluations of standard models~\cite{mujtaba-etal-2024-lost,Mitra2021AnalysisAT} have shown that mainstream ASR systems do not perform well in stuttered speech. Beyond these technical gaps, prior work has also been limited by a lack of in-depth stakeholder collaboration. While researchers often engage with end users (i.e., \slp and \pws) for data collection and annotation, we find that relatively few studies report end user participation in problem formulation, task definition, annotation design, evaluation, or interpretation of system errors. Yet, end users are uniquely positioned to align model outputs with decision contexts, anticipate deployment constraints and harms, and articulate what ``\emph{useful}'' actually means in practice. Without systematically incorporating these points of view, it is difficult to ensure clinical relevance, problem–solution fit, ecological validity, and transferability to the real-world, or to prioritize requirements for a truly human-centred design~\cite{Li2025OurCV,Sridhar2025JjjjustSB}.

\begin{figure}[t]
    \centering
    \includesvg[width=1\linewidth]{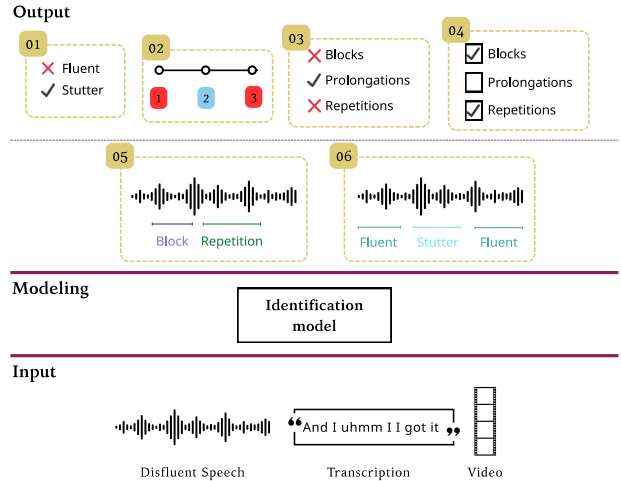}
    \caption{Variation in inputs, task formulation and outputs for the stutter identification research area. We propose a task taxonomy (see Section \ref{sec:ResearchAreas}) to handle variation in output forms and standardise research nomenclature.}
    \label{fig:identification}
    \vspace{-1.5em}
\end{figure}

Several recent studies highlight both the need and the willingness of stakeholders to engage. Colin et al.~\cite{Lea2023FromUP} surveyed perceptions of \pws about speech recognition systems and provided recommendations for building better user-centric tools, specifically for voice-assistant systems. Jingjin et al. \cite{Li2025GovernWN} conducted a survey involving the stuttering community about data governance preferences for speech AI and derived best practices for ethical data collection, sharing, and use, underscoring that the community is both engaged and opinionated about how speech technology should be built. These works focus on \pws~perception and needs from specific speech technologies (voice assistants, data governance). Existing survey-based work on stuttered-speech \cite{10.1016/j.neucom.2022.10.015, Khara2018ACS} tends to focus on particular technologies or design questions rather than on the field as a whole. In contrast, we take a broader view: we analyse how the overall stuttered-speech research agenda aligns with the experience and needs of the main end-users of the resulting speech technology. 

%: \pws and \slp.

We argue that many of the current gaps stem from limited stakeholder involvement in problem definition and model evaluation. In ASR work, what counts as a “correct’’ output is usually defined by generic benchmarks rather than by clearly specified, user-centred use cases. For example, \slp prefer transcripts that are true to produced speech for assessment and documentation, while \pws favour transcripts that represent intended speech for communication support. Similarly, ambiguities can arise in research output stemming from  variations in task naming and formulation - for example, between stutter \emph{detection} and \emph{classification}, and in the level of output granularity (shown in  Figure \ref{fig:identification}). Without standardised task names and definitions, it is hard for the research community and stakeholders to know which output format a given system is optimising for, or to design benchmarks that separately and fairly evaluate different objectives. We therefore propose a task taxonomy that makes these distinctions explicit, so that user-centred objectives are targeted more precisely and measured more effectively in future research.

In this paper, we focus on two primary stakeholder groups: \pws~and \slp, and we ask four questions:

\begin{enumerate}
        \item \textit{What research areas do current stuttered-speech \ztedit{research} papers prioritise?}
         \item \textit{What do \pws~and \slp~say they need from voice-based \ztedit{technology}?}
         \item \textit{To what degree do current research practices align or conflict with stakeholder needs and priorities?}
         \item \textit{What are the implications of these findings for future stuttered speech technology research?} 
\end{enumerate}

To answer these questions, we combine evidence from both the literature and stakeholders. First, we map $228$ stuttered-speech \ztedit{research} papers using our proposed task taxonomy and annotated them for language coverage, stakeholder collaboration, and open-science practices. Second, we establish stakeholder needs, priorities and pain points through survey results from $40$ \pws and $30$ \slp. Third, we analyse the alignment between the current state of stuttered-speech research and stakeholders' responses, identifying systematic gaps, opportunities, and implications. Finally, we translate our findings into \ztedit{concrete} recommendations for future research directions, evaluation practices, and interdisciplinary protocols aimed at making stuttered-speech research truly user-centred.

\textbf{Paper Structure:}
The remainder of the paper is structured as follows. 
Section~\ref{sec:literatureFindings} maps the current research landscape for stuttered-speech technology, introducing our task taxonomy and summarising key patterns in these research areas. %, language coverage, open-science, and stakeholder involvement. 
Section~\ref{sec: surveyDesign} describes the design and methodology of our stakeholder surveys, while
section~\ref{sec:stakeholder needs} presents our analysis of stakeholder insights and needs related to speech technology. %, detailing \pws and \slp experience with existing tools and what they say they need from voice-based AI. 
In Section~\ref{sec:alignment}, we bring these strands together % by analysing how they align,
by contrasting research with stakeholder priorities and highlighting systematic gaps and design implications. 
Finally, Section~\ref{sec:conclusion} outlines our actionable recommendations for future research. % directions, evaluation, and interdisciplinary collaboration.

\section{Literature Mapping}
\label{sec:literatureFindings}
% \ZC{the previous bit could come here, with some additional text to introduce it.}

\subsection{Search and Annotation Protocol}
% \ZC{this next bit might read unclear, so feel free to restore to previous versions/edit freely.}
We conducted a scoping review of research work on stuttered speech processing to characterise the current state of research. 
% using the Semantic Scholar API.\footnote{\url{https://www.semanticscholar.org/product/api}}
\ztedit{To identify relevant papers, we performed a keyword search resulting in $680$ papers published between 2010 and October 2025 across all open access publications indexed by Semantic Scholar.\footnote{Our search terms were: \dq{disfluency/dysfluency}, \dq{stuttering/stutter}, \dq{recognition}, \dq{detection}, \dq{classification},\dq{artificial intelligence}, \dq{machine learning}, \dq{language model}.}}
\ztdelete{Their keyword-based paper selection method first yielded approximately $680$ papers from 2010 - October 2025 using the terms:  \dq{disfluency/dysfluency}, \dq{stuttering/stutter}, \dq{recognition}, \dq{detection}, \dq{classification},\dq{artificial intelligence}, \dq{machine learning}, \dq{language model}.} 
\ztedit{We then reviewed papers on the basis of their titles and abstracts, excluding all papers whose primary focus was clinical, neurological or behavioural and did not have a substantial machine learning or speech technological component. 
In our full reading of the remaining $234$ papers, we excluded a small number of papers whose content did not match the abstract or were proposals rather than completed empirical work, resulting in a final corpus of $228$ papers. We did not restrict inclusion by publication venue (e.g., major speech conferences%\footnote{Interspeech, ICASSP, ASRU, IEEE SLT}
); doing so would have removed more than 70\% of otherwise eligible papers. List of final papers is publicly available.\footnote{\label{fn:metalink}\scriptsize\url{https://anonymous.4open.science/r/stutterresearch_survey-D783}}
%All remaining papers were manually annotated along four main dimensions:
}
%without a substantial AI or speech-technology component, since our goal is to characterise how AI research addresses stuttered speech to enable researchers and stakeholders to ``speak together''.}
%
\ztdelete{After title and abstract screening, we excluded non-open-access works and papers whose primary focus was clinical, neurological or behavioural without a substantial AI or speech-technology component, since our goal is to characterise how AI research addresses stuttered speech to enable researchers and stakeholders to “speak together”. This left $234$ papers. Full-text reading led us to discard a small number of works whose main content did not match the abstract or that read more like proposals than completed empirical work, resulting in a \textbf{final corpus of $228$ papers}. We did not restrict inclusion by publication venue (e.g., major speech conferences); 
%\footnote{Interspeech, ICASSP, ASRU, IEEE SLT}
doing so would have removed more than 70\% of otherwise eligible papers. Each paper was read in full and \textit{manually} annotated along four main dimensions:}

\subsection{Annotation Guidelines}

% \ztedit{}
We manually annotated our final sample of $228$ papers according to four main dimensions: research areas, language coverage, stakeholder involvement, and open-source availability. To ensure consistent coding of the literature and enable systematic comparison with stakeholder needs, we first developed a common vocabulary of research areas through an iterative labelling and clustering process. In the following sections, we describe this vocabulary by giving clear definitions of the principal research areas identified in the stuttered-speech literature. For some of these areas, we additionally define \emph{sub-tasks} to capture important variations in research focus that we later use in the alignment analysis.

% To ensure consistent coding of the literature and enable systematic comparison with stakeholder needs, a \emph{common vocabulary} is required. This subsection also provides\hanan{This is mixing the annotation methodology with the findings; you can describe instead how the common vocabulary was found, and then state that the remaining section describe these findings. } clear definitions of the principal research areas identified within the literature on stuttered speech. For some of these areas, the further defined \emph{sub-tasks}  to capture variations in research focus for subsequent alignment analysis.
\begin{figure*}[t]
        \centering
        \includegraphics[width=\linewidth]{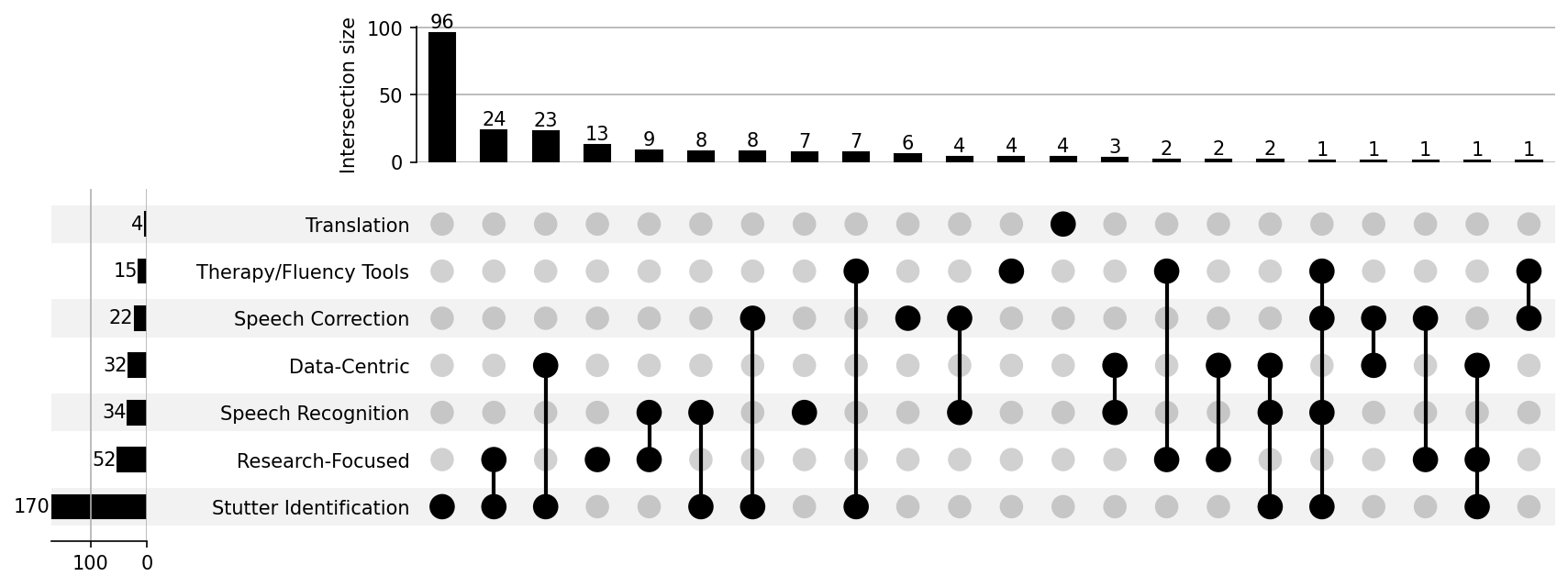}
        \caption{Task Combinations in Literature Survey. The left bar chart shows distribution of frequency of research areas in the survey. The top bar chart shows the frequency of combination of research areas.}
        \label{fig:taskCombination}
        \vspace{-1.5em}
\end{figure*}
\subsubsection{Proposed Research Area Taxonomy} 
\label{sec:ResearchAreas}

\begin{table}[h]
\centering
\small
\setlength{\tabcolsep}{6pt}
\caption{A taxonomy of identified research areas and sub-tasks in stuttered speech research and example papers.}
\renewcommand{\arraystretch}{1.2}
\resizebox{\columnwidth}{!}{%
\begin{tabular}{@{}lll@{}}
\toprule
\textbf{Research Area} & \textbf{Sub-Tasks} & \textbf{Example Papers} \\
\midrule

\multirow{2}{*}{Speech Recognition} 
& Intended Speech Recognition & \cite{Romana2024FluencyBankTA} \\
& Verbatim Speech Recognition & \cite{Mujtaba2024InclusiveAF,Mendelev2020ImprovedRT}\\

\midrule

\multirow{3}{*}{Stutter Identification}
& Stutter Classification & \cite{Miyahara2025StutteringDB, Kommagouni2025TowardsCO} \\
& Stutter Detection & \cite{Zhou2024StutterSolverEM}\\
& Stuttering Severity Assessment & \cite{valente25_interspeech}\\
\midrule

\multirow{1}{*}{Translation}
& -- & \cite{Salesky2019FluentTF,Salesky2018TowardsFT,Saini2020GeneratingFT}\\

\midrule

\multirow{1}{*}{Speech Correction}
& -- & \cite{Murugan2022EfficientRA, Rajput2021SpeechSD,2020AutomaticSR,Bhat2023AdversarialTF,Bhat2023DISCOAL} \\

\midrule

\multirow{1}{*}{Therapy / Fluency Tools}
& -- & \cite{Faggiani2025DEMOEA,Vona2023SpeakIP,Heeman2011ComputerAssistedDC, Heeman2016UsingCA}\\

\midrule

\multirow{2}{*}{Data-centric}
& Resource Curation & \cite{Batra2025BoliAD,bayerl-etal-2022-ksof, Romana2024FluencyBankTA}\\
& Synthetic & \cite{zhang25u_interspeech, Kourkounakis2021FluentNetED,Sen2021SemanticPO}\\

\midrule

\multirow{2}{*}{Research-focused} 
& -- & \cite{Hintz2023AnonymizationOS,Wong2024DistillingDU,kouzelis23_interspeech,10.1016/j.neucom.2022.10.015,Khara2018ACS,Evangeline2024InvestigatingAA}\\
& Ethics & \cite{Li2025OurCV, Li2025GovernWN}\\

\bottomrule 
\end{tabular}
}
\label{tab:stutter_taxonomy}
\vspace{-1em}
\end{table}

\ztedit{
We perform an inductive coding of research areas, assigning each paper one or more research areas according to its contributions (see Table~\ref{tab:stutter_taxonomy} for an overview of research areas).} Under some research areas, we introduce \emph{sub-tasks}, which describe variations within a research area that are either often conflated in literature, or required for in-depth alignment analysis. 

% \red{introduce sub-task}

\vspace{0.2cm}
\noindent\textbf{Speech Recognition}
\ztedit{
First, we identify Stuttered Speech Recognition as a primary research area, defined as the transcription of stuttered speech audio into text. 
We further distinguish between \textit{{intended}} and \textit{{verbatim}} speech recognition. 
\textit{\underline{Intended speech recognition}} aims to remove all stuttering events from the transcription to convey the intended message more effectively, whereas \textit{\underline{verbatim speech recognition}} aims to remain faithful to the produced speech, including repeated words and syllables, filler words, or interjections produced as a result of stuttering.
}

\vspace{0.2cm}
\noindent\textbf{Stutter Identification}
Second, we identify stutter identification as a primary area, which aims to analyse the occurrence and/or intensity of stuttering events in produced speech. 
We further divide this research area into \textit{\underline{stutter/disfluency classification}}---which seeks to map pre-segmented  audio or audiovisual clips into event types, using binary classification, multi-class classification, or multi-label classification ($01$, $03$, and $04$ in Figure~\ref{fig:identification}, respectively)\footnote{We note that ``stuttering'' and ``disfluency'' are used interchangeably in papers; however, stuttering is a multi-dimensional disorder that is not limited to verbal disfluencies but can also include non-verbal behaviours essential for accurate diagnosis \cite{valente25_interspeech}.};  \textit{\underline{stutter/disfluency detection}}---which seeks to identify the time bounds of stuttering or disfluency events in an audio or audiovisual clip, then map these segments to event types (see $05$ and $06$ in Figure~\ref{fig:identification})\footnote{A notable issue in the literature is the inconsistent use of `classification' and `detection', which are often used interchangeably, despite referring to distinct problem formulations and output structures.}; and \textit{\underline{stuttering severity assessment}}---which seeks to assign a  severity level to an audio or audiovisual sample, either as a regression task assigning a value on a numerical scale, a binary classification task, or a multi-class classification task (see $02$ in Figure~\ref{fig:identification}).

\vspace{0.2cm}
\noindent\textbf{Others}
We identify a series of other smaller research areas that we categorize as follows: \textit{Translation}, capturing methods for producing fluent translations from either disfluent speech or transcripts; \textit{Speech Correction}, aiming to remove stuttering events to generate fluent speech from disfluent inputs; \textit{Therapy and Fluency Tools}, which describe frameworks or applications for stakeholders, such as fluency practice or  tools for assessment and diagnosis; \textit{Data-Centric Approaches}, referring to efforts towards resource development, including data collection and curation efforts (e.g., offering new speech data or annotating existing corpora) and data augmentation efforts such as \textit{synthesizing} disfluent speech. 
Finally, we capture remaining works within the \textit{Research-Centric} category, which includes works that %focused on improving research outputs for stuttered speech, through works
(1) contribute to technical tasks or meta studies (e.g., forced alignment, uncertainty estimation, and literature reviews) and (2) works that focus on ethics, such as identifying biases in research for stuttered speech or describing user perspectives on speech technologies for stuttering.

\subsubsection{Discussion}
The main conceptual novelty in our proposed taxonomy lies in disentangling two heavily confounded parts of the field: \emph{speech recognition} for stuttered speech and \emph{stutter identification}, where methodological variability, inconsistent task formulation and ambiguous naming are common in the literature. The research area names defined in \emph{Others} are primarily used to thematically code the literature. Unlike the sub-task distinctions introduced for speech recognition and stutter identification, these areas are not especially ambiguous in current work; they simply provide a consistent way to group papers into broad categories. 

\ztdelete{
\begin{itemize}
    \item \textbf{Research areas}: we assigned one or more labels from taxonomies discussed in section~\ref{sec:ResearchAreas} to each paper, allowing multiple labels whenever a work made a substantive contribution in more than one area. For example, a paper presenting a new method for stutter event classification and also proposing a novel data augmentation strategy is labelled \emph{Stutter Identification} and \emph{Data-centric}.
    \item \textbf{Language coverage}: we record all languages and dialects reported in the paper, as used for training or evaluation. For works without explicit language specification, we label as \emph{unclear}, especially if implicit identification (e.g., from dataset) is not possible.
    \item \textbf{Stakeholder involvement}: we noted whether and how \pws and \slp were involved beyond data provision and/or annotation. For example, in study design, evaluation, clinical testing or providing expert opinion.
    \item \textbf{Open-source availability}: we marked whether the paper released datasets, trained models, or code sufficient for reproducing the main experiments.
    % \item \textbf{Summary}: we noted interesting insights from the papers.
\end{itemize}
}

\ztdelete{
\subsection{Research Areas in Literature}
\label{sec:ResearchAreas}
To ensure consistent coding of the literature and enable systematic comparison with stakeholder needs, a \emph{common vocabulary} is required. This section therefore provides clear definitions of the principal research areas identified within the literature on AI for stuttered speech. These research areas were derived through an iterative reduction process. Initially, each study was assigned a descriptive phrase summarizing its primary contribution (e.g., ``time-wise stutter detection on read speech'' or ``synthetic stuttering data generation''). These phrases were then grouped using thematic coding. Overlapping or conceptually similar groups were progressively merged until a stable taxonomy of seven distinct research areas was established. The main conceptual novelty lies in disentangling two heavily confounded parts of the field: speech recognition for stuttered speech and stutter identification, where methodological variability, inconsistent task formulation and ambiguous naming is common in literature. Table \ref{tab:stutter_taxonomy} summarizes the research areas and sub-tasks with some example papers.
}

\ztdelete{    
\subsubsection{Proposed Sub-Task Taxonomy}
\label{sec: Taxonomy}
\noindent\textbf{Speech Recognition.} Defined as a speech-to-text problem in literature, where the desired output is transcriptions of spoken speech.  We describe $2$ sub-tasks for speech recognition:  
\begin{itemize}
    \item \textit{Intended speech recognition}: here, the transcription is true to the intent of the speaker, with produced stuttering events completely removed from the transcripts. A use-case for this sub task is in voice-activity commands or dictation tools. In research, when evaluating the machine-generated transcription of a read speech that includes stuttering moments, the reference transcription should be the read passage. 
    \item \textit{Verbatim speech recognition}: in this form, the transcription is true to the produced speech, where the repeated syllables, filler words or interjections produced as a result of stuttering are included in the machine-generated transcription \textit{(e.g. ``I $\backslash$r won'' \textbf{or} ``I $\prec$rep$\succ$ won'' \textbf{or} ```I I I won'')}. \footnote{In this example $\backslash$r and $\prec$rep$\succ$ stands for repetition stutter event.}
\end{itemize}
These two sub-tasks align with different stakeholder needs: \pws typically want intended transcripts for everyday communication, whereas \slp often need verbatim transcripts for assessment and documentation.
}

% \vspace{0.5em}
\ztdelete{
\noindent\textbf{Stutter Identification.} Involves determining the occurrence/intensity of stuttering events in produced speech. This task has a high degree of variability in terms of sub-tasks according to literature. The variability occurs from the level of granularity of model inputs, outputs, etc. as shown in Figure \ref{fig:identification}. We describe 3 sub-tasks for stuttering identification that clarify the expected output: 
\begin{itemize}
    \item \textit{Stutter/disfluency Classification} refers to a mapping problem from audio clips to classes. In this sub-task there is some variability in the number of \textit{class} mappings: binary classification (\emph{01 in Figure \ref{fig:identification}}), multi class classification (\emph{03 in Figure \ref{fig:identification}}) and, multi label classification (one audio clip is mapped to one or more stuttering events (\emph{04 in Figure \ref{fig:identification}}).
    \item \textit{Stutter/disfluency Detection} describes time-frame to class mapping. Differs from classification in that identification is more fine-grained to the temporal (seconds/frame) occurrence level of the event. Given an audio clip, the task formulation involves identifying time-bound occurrences of stuttering events, sometimes at a binary level and other times at a multi-class level (\emph{05 and 06 in Figure \ref{fig:identification}})
    \item \textit{Stuttering Severity Assessment}: this refers to the task of assigning a severity label to an audio clip. The task could be framed as a regression (on a numerical scale, \emph{05 and 06 in Figure \ref{fig:identification}}) or a multi-class problem (mid, low, high). All the subtasks described for identification can be framed in a multi-modal format, using textual and/or visual information for modelling.
\end{itemize}
A notable issue in literature is the inconsistent use of terminology, i.e., the terms \emph{detection} and \emph{classification} are frequently used interchangeably, despite corresponding to distinct problem formulations and output structures. In many cases, studies labelled as “detection” in fact perform clip-level classification. In addition, the terms ``disfluency'' and ``stuttering'' are often used interchangeably in the literature, but stuttering is a multi-dimensional disorder that includes but is not limited to verbal disfluencies.
% \hanan{I added this second part; please check.}
}
\ztdelete{
\subsubsection{Other Research Areas}
The research area names defined here are primarily used to thematically code the literature. Unlike the task distinctions introduced earlier (Section~\ref{sec: Taxonomy}), these areas are not especially ambiguous in current work; they simply provide a consistent way to group papers into broad categories. For some areas, we further define sub-tasks to capture variations in research focus for subsequent alignment analysis.
\begin{enumerate}
    \item \textbf{Speech correction}: describes attempts to remove stutter events, typically done at speech (generating fluent speech from atypical speech) or transcript level (removing disfluent events in transcripts).
    \item \textbf{Data-centric}: refers to efforts towards resource development and data augmentation. This includes: (i) \emph{Resource Curation:} efforts to either produce real recordings of \pws or provide updated annotations to existing corpora to benchmark new tasks, and (ii) \emph{Synthetic:} works that propose data augmentation strategies and methods for synthetic disfluent speech generation.
    \item \textbf{Translation}: describes research efforts on methods for producing fluent translations from either disfluent speech or disfluent transcription.
    \item \textbf{Research-focused} There are a number of works in the literature that are not application-focused, they are more towards improving research outputs for stuttered speech including: (i) Speech processing works on forced alignment, uncertainty quantification, evaluations and ablation studies or surveys, and (ii) \textit{Ethics:} work on the intersection of research and humans, involving  identification of biases in research for stuttered speech or describing user perspectives on AI tools.
    \item \textbf{Therapy / Fluency Tools}: refers to works/demos that propose frameworks or applications for \pws and~\slp. This includes tools built for assessment or identification of stuttering, and tools built for fluency practice.
\end{enumerate}
}

\subsection{Findings: State of the Literature}
\label{sec:literatureCount}

% \subsubsection{Research Area Combinations.} 
% \label{sec:intersection}

% \ZC{I don't understand this notion of combination and intersection. Intersection/combination of what? Different research areas? \response{H} This referes to combination of research areas in a paper. \response{Z} Okay cool, can you just make that clear?}
\ztdelete{Figure~\ref{fig:taskCombination} shows that stuttered-speech AI research is overwhelmingly dominated by \textbf{stutter identification}, which appears in \textbf{170 papers}.} 
We show frequency of combinations and occurrence of research areas in the literature in Figure~\ref{fig:taskCombination}. 
\ztedit{Research in stuttered-speech processing is dominated by \emph{stutter identification} (n=$170$).
}The single largest group (\ztedit{n=}$96$\ztdelete{ papers}) focuses exclusively on stutter identification. 
The next most common intersection of works combine stutter identification with \emph{research-focused} (n=$24$) and with \emph{data-centric} work (n=$23$), after which intersection sizes drop rapidly. 
Most combinations occur in fewer than $13$ papers, with many (4 combinations in total) appearing only once. 
\emph{Therapy/fluency tools} are intuitively most often developed with stutter identification models (n=$7$).

\subsubsection{Sub-task Frequency and Practices.}
% \ZC{I removed the framing of sub-tasks so we might need to update this (or return the subtask framing).}
% \red{ToDo: figure or table for count of papers in sub-tasks}
Research on stutter identification is heavily dominated by the classification sub-task: either performing \textbf{binary classification} ($38$ papers), \textbf{multi-class classification} with \textit{widely varying label sets} (typically 3–8 classes; sometimes tied to specific event taxonomies such as repetition types; $88$ papers), or \textbf{multi-label classification} ($7$ papers). \dq{Detection} and \dq{classification} are often used interchangeably for disfluency identification work, and titles rarely make the actual task clear. Although $72$ papers include \dq{detection} in their title, only $6$ of these perform a temporal detection task under our definition; the others perform some form of classification on pre-segmented inputs (typically 3-5 seconds long audio). In addition, most works focus exclusively on verbal disfluencies, which are the most common types of stutter events, but clinical definitions distinguish between verbal disfluencies and other types of stuttering events \cite{valente25_interspeech}. This distinction is rarely acknowledged in the surveyed literature. These omissions and naming inconsistencies make it difficult to compare works or identify relevant methods for specific downstream tasks. Standardized task definitions would make it easier to identify and compare related literature.

For stuttered speech recognition tasks, intended speech recognition is the \emph{implicit standard} in the literature. Most works use evaluation metrics, references, and design choices (e.g., filler word removal) that implicitly optimize for intended speech recognition without explicitly acknowledging this choice.
We often inferred the specific ASR sub-task through dataset labels or down-stream task. In $\sim7\%$ of the primarily stutter ASR works, we were unable to determine if verbatim/intended ASR was the objective. About $59\%$ of ASR works focus on intended ASR, while the rest focus on verbatim ASR. Recently, verbatim ASR has been framed mainly as a form of identifying disfluencies, where special tokens are used to represent disfluencies \textit{(e.g., models would output ``I $\backslash$r won'' \textbf{or} ``I $\prec$rep$\succ$ won'' \textbf{instead of} ``I I I won'')}.\footnote{In this example $\backslash$r and $\prec$rep$\succ$ stands for repetition stutter event.} %About $90\%$ of works from 2023. \hanan{This isn't clear. 90% of which works? }

Within the research-focused area, there's an uneven mix of sub tasks: about $50\%$ of the papers are on surveys/evaluation of research methods, $\sim12\%$ examine ethical aspects of stuttered-speech research, sometimes involving \pws~perspectives, $\sim4\%$ (n=$2$) explore anonymization and privacy for stuttered-speech research, $\sim4\%$ propose approaches towards language generalization, and $\sim4\%$  examined explainable approaches for stutter classification. These numbers motivate our later analysis of research-focused gaps in Section \ref{sec:alignment}.

\subsubsection{Language Coverage.}
\begin{figure}[h]
    \centering
    \includegraphics[width=1\linewidth]{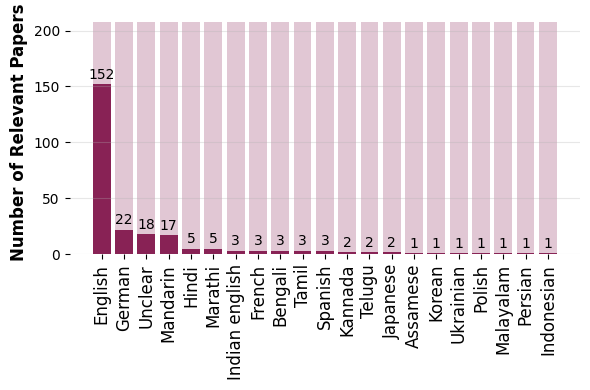}
    \caption{Languages covered in literature shows stuttered-speech AI research is English-centric. The long-tail indicates notable but minimal efforts towards language diversity.}
    \label{fig:langDist}
    \vspace{-1.em}
\end{figure}

The language distribution in the stuttered-speech \ztdelete{AI}\ztedit{technology} literature is highly skewed towards English \ztedit{with 152 papers focusing on English}.
\ztedit{In contrast, the next four most common languages are German (22 papers), Mandarin (17 papers), Hindi (5 papers), and Marathi (5 papers). All other languages in our sample are covered by three or fewer papers (see Figure~\ref{fig:langDist}).} This indicates that most modelling, benchmarking, and evaluation choices are implicitly optimized for high-resource languages. There were instances (n=$18$) of incomplete documentation, where experiment language is not mentioned explicitly and cannot be inferred from data resource used. The majority of studies are monolingual (n=$183$), with relatively few multilingual (n=$25$) efforts. There's relatively limited work ($n<10$) on cross-lingual transfer, language-agnostic representations, or systems designed to operate robustly across languages.
% The current monolingual English-centric state of research makes it difficult for models to be adopted and trusted by communities whose languages and linguistic realities are largely absent from current research. 
% \hanan{The focus on low-resource languages at the end of these 2 subsections feels rather off-topic. The paper already feels like it's trying to do too much, so not sure these additional discussions on fairness fit the main narrative.}

% \ZC{add something about the implications. I've heard from someone in the calls, I think, that stuttering has variety across languages---if this is true then we should say that clearly.} 
% \zc{add a "so what" sentence here. Why does this matter?}
% \ztdelete{English accounts for 152 papers, \ztdelete{far exceeding the next most common languages:} German (22), and Mandarin (17) and a long tail of other languages (see Figure \ref{fig:langDist}).} 

\subsubsection{Collaboration, Open-Science and Trends.}
\begin{figure}[h]
    \centering
    \includesvg[width=0.86\linewidth]{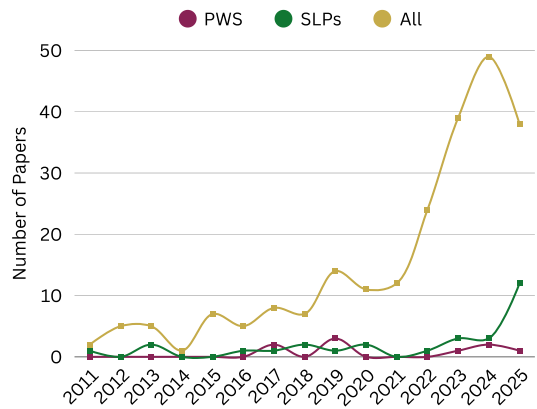}
    \caption{Number of papers published by year. PWS means papers involving \pws collaboration; SLPs means papers involving \slp collaboration; All means all papers published.}
    \label{fig:stakeholder_involvement}
    \vspace{-1em}
\end{figure}
Fewer than 20\% of papers explicitly report inter-disciplinary collaboration with stakeholders, indicating that much of the research area combinations remain more technically focused rather than truly human-centred.
% \zc{this brings back that I don't think I understand section \ref{sec:intersection} or our findings in it.} \ztdelete{Stakeholders are collaborating in these ways in the literature:}
\ztedit{Stakeholders are collaborating in work around } data curation (n=$24$), interview/user study respondents (n=$2$), expert opinions (n=$3$), evaluation (n=$7$) and clinical testing (n=$3$). A noticeable trend is the increase in works that report stakeholder involvement (Figure \ref{fig:stakeholder_involvement}), even if they are majorly towards data curation.

% \zc{Be more specific and precise here. What does rise in stakeholder collaboration mean?} 

% It is striking that the most studied task area (stutter identification) has so few

There's a recent, albeit small, rise in open-science practices for stuttered-speech \ztedit{systems}. Only $23$ papers ($\sim10\%$) report releasing any combination of code, models, or data starting from 2021, coinciding with rise in the number of published papers (see Figure \ref{fig:stakeholder_involvement}). Among these: $12$ are data-centric, $5$ focus on speech recognition, $4$ focus on stutter identification, and $2$ are research-focused. Only around $8\%$ of works that propose stutter identification (\emph{most common research area}) provide open-source resources. 

We also observed a growing shift towards high-resource frameworks such as large synthetic datasets ($28\%$ of all data-centric papers; first occurrence in 2021), large-scale pre-trained models, or pipelines that depend on extensive lexical resources.

% The papers that support open-science all fall in the year range of 2021 - 2025. 
% From 2021, we see a significant rise in the number of published papers (see Figure \ref{fig:stakeholder_involvement}). 

% , which constrains reproducibility and downstream user adoption

% \zc{rephrase this. Maybe specifically say the percentage of papers that do open methods for stutter identification specifically.}

% \subsubsection{Research Trends}

% The term \textit{detection} and \textit{classification} are often used interchangeably, and titles rarely make the underlying task formulation explicit. For example, although \textbf{72 papers} include \textit{\dq{detection}} in their title, only \textbf{6} actually carry out a detection task under our definition.
\ztedit{\subsubsection{Discussion}}
Overall, our findings indicate that the stuttered-speech technologies field is active, but narrowly scoped. The literature is overwhelmingly organized around monolingual stutter classification tasks, with inconsistent task naming and an implicit default toward intended speech recognition. Stakeholder collaboration and open-science practices are still the exception rather than the rule, and work that investigates ethics, explainability, or deployment remains sparse. The limited amount of open source works reduces opportunities for incremental progress in adapting tools to local clinical and diverse linguistic contexts in the research community. In practice, limited openness further weakens the potential impact of research on the stakeholder communities it aims to support.  Recent trends give encouraging signs: more papers are being published in the area, there is a gradual rise in stakeholder involvement, and open datasets and synthetic-data efforts, suggesting growing awareness of some of the field’s underlying limitations. However, addressing these methodological issues in isolation does not guarantee that systems will serve \pws~and \slp~well: some gaps are starting to close, but the extent to which the current state of research actually aligns with their needs remains an open question.

% \zc{Have a more punchy line to end this section with.} 

% \zc{I think you can cut this last line altogether. In the beginning of the next section you just need to make clear that the shift is happening. Also fine to keep it, but in that case you should make the transition smoother.}

% In the next sections, we compare the current state of the stuttered-speech literature, as characterized in this section, with the needs expressed by \pws and \slp, as outlined in Section \ref{sec:stakeholder needs}. We then revisit these findings in a dedicated alignment analysis presented in Section \ref{sec:alignment}, where we systematically examine the extent to which existing research practices correspond to identified stakeholder priorities.

\section{Stakeholder Survey}
\label{sec: surveyDesign}

\subsection{Survey Methodology}
One of the main goals of this work is to characterise what stakeholders' expectations are and identify how stuttered-speech research can better serve stakeholders. We designed two complementary online questionnaires: one targeting \pws~and one targeting \slp. Both questionnaires were co-designed iteratively with two experienced \slp, in English and Spanish. The language choice followed the \slp~recommendation. The \slp first drafted clinically grounded questions based on their practice and research experience. We then added items capturing stuttered speech technology-related perspectives, such as the perceived usefulness of different tool types. We received responses from \textbf{40} \pws and \textbf{30} \slp. \\
% and willingness to participate in data initiatives

\noindent \textit{\textbf{Persons Who Stutter (\pws) questionnaire:}} This survey consisted of $35$ questions across four sections:
\begin{itemize}

    \item \textbf{Demographic and background:} age, gender, country of residence and origin, languages spoken with fluency levels, education, and employment status.

    \item \textbf{Stuttering profile:} impact on professional activity, self-reported severity and frequency, differences in severity across spoken languages, and how severity correlates with specific situations or people.

    \item \textbf{AI tool utility measure:} perceived usefulness of AI tools in participants’ daily lives, and willingness to take part in research-focused activities.

    \item \textbf{Needs and perceptions of voice-based AI Tools:} current usage and familiarity, challenges and concerns, and recommendations for future research and tool development.
\end{itemize}

\  \\

\noindent \textit{\textbf{Speech-Language Pathologist (\slp) questionnaire:}} The survey consisted of 38 questions across four sections:
%, addressing respondents’ demographics and professional profile, current work life and challenges, the perceived utility of AI tools, and current AI tool use. 

\begin{itemize}

    \item \textbf{Demographics and professional profile:} 
    age, gender, country of practice, clinical practice languages with fluency levels, perceived similarity of stuttering characteristics across languages and age groups, education level, years of experience, and frequency of working with \pws.
    
    \item \textbf{Current work life and challenges:} difficulties encountered when working with \pws, challenges in day-to-day clinical practice, and willingness to discuss these challenges with researchers.
    
    \item \textbf{AI tool utility measure:} usefulness of literature-identified tools for clinical work with \pws, importance and stage of interdisciplinary collaboration, and importance of tool generalizability across languages.
    
    \item \textbf{Use of AI tools:} current usage and familiarity, perceived benefits, challenges and concerns, field readiness and confidence in adopting AI tools, and needs and recommendations for developing and integrating AI tools.
\end{itemize}

\  \\

The surveys were distributed via the professional and clinical networks of the collaborating \slp and through their local communities over a 3-month period. Participation was voluntary and uncompensated; respondents gave informed consent for analysis and reporting of aggregate results.

The questions\footref{fn:metalink} used a combination of five-point Likert scales, multiple-choice questions often with optional open-text fields for elaboration or additional suggestions, and open-ended text input. Quantitative responses were summarised using descriptive statistics (e.g., counts and proportions). Free-text responses were thematically coded to identify recurring needs, concerns, and design recommendations. This method allows us to understand “\textit{what gets researched}” and to compare this with each stakeholders' views of “\textit{what is needed and experienced}” in everyday communication for stuttered speech.

\section{Stakeholder Insights}%: The Human Reality}
\label{sec:stakeholder needs}
This section highlights the needs of \pws~and \slp from stuttered-speech technologies by analysing their responses to the corresponding questionnaires. 
We include a small number of anonymised, verbatim excerpts from open-text survey responses to illustrate recurring themes and to contextualise the quantitative results. Quotes are labelled by respondent group (\pws{} or \slp{}).

\subsection{Respondents, Variability and Everyday Impact}

\begin{figure}[ht]
    \centering
    \includesvg[width=\linewidth]{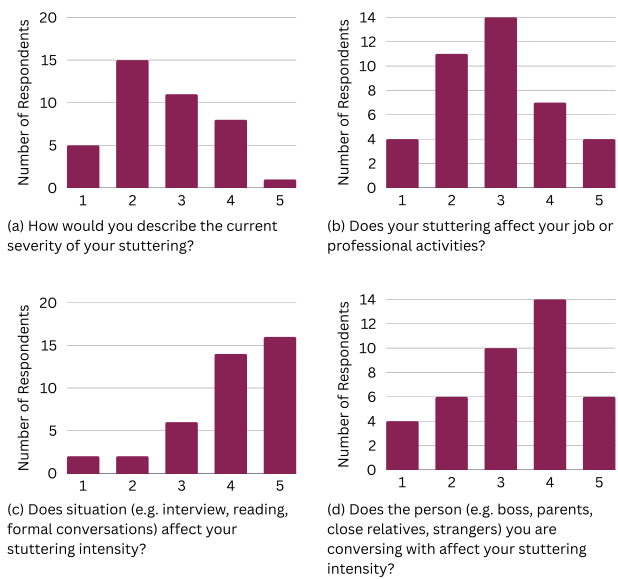}
    \caption{\pws responses to Likert-scale questions on stuttering variability.}
    \label{fig:pws_variability}
    % \vspace{-1.em}
\end{figure}
% \begin{figure}[h]
%     \centering
%     \includesvg[width=1\linewidth]{images/PWS_demo}
%     \caption{Gender and Age distribution of PWS Respondents.}
%     \label{fig:pws_demo}
% \end{figure}
\pws respondents formed a young and multilingual group. Most were between 18 and 44 years old, and more than $90\%$ reported speaking at least two languages; around one-third spoke three or more. On a five-point scale, most \pws rated their current stuttering severity as mild to moderate (levels 2–3; Figure~\ref{fig:pws_variability}(a)). The responses mostly clustered around the middle of the scale when being asked about the impact on work (Figure~\ref{fig:pws_variability}(b)) indicating that stuttering moderately affects their job or professional activities. Many described noticeable variation in their stuttering across languages (native vs. additional languages), situations (e.g., interviews, reading aloud, formal conversations) and conversational partners (e.g., work supervisors or colleagues, family members, strangers). Importantly, these averages obscure strong context effects: most \pws reported substantial spikes in their stuttering intensity in high-pressure or formal situations and with certain interlocutors (Figure~\ref{fig:pws_variability}(c,d)). This suggests that \pws burden is often driven less by a stable severity level and more by context-dependent breakdowns in fluency—precisely the scenarios where voice-based systems and clinical monitoring tools must remain reliable. Consequently, evaluations that only report a single global severity or an overall performance score risk underestimating real-world difficulty and may fail to reflect the situations that matter most to people who stutter.
%
% The \slp respondents were highly experienced. 

Two-thirds of \slp~reported more than 10 years of clinical practice, and a large subset (10/30) had more than 20 years. Experience working with \pws specifically was more evenly distributed: while over half (16/30) had 0–5 years of \pws-focused experience, the others reported at least 6 years. In addition, approximately one-third reported working in multilingual contexts, most commonly with $2$ languages. This suggests the variability in specialist exposure and multilingual complexity, highlighting the need for tools that remain robust across languages and service contexts to complement their expertise.

% are easy to integrate into routine practice and 

\slp and \pws engage with stuttering in different roles: clinical documentation/treatment vs. everyday communication. When asked about challenges that they came across, their responses were complementary.
\slp highlighted challenges such as the time required for detailed transcription and analysis, difficulties in tracking stuttering variability across contexts and over time, and the lack of automated tools to support documentation, severity assessment and longitudinal monitoring. In contrast, \pws reported challenges primarily related to real-time communication, self-monitoring, voice-assistant support and emotional regulation. 

\subsection{``Impatient ASR'' and Divergent Transcription Needs}
\pws experiences with existing voice-based AI tools highlight what we call the ``\textit{Impatient ASR}'' problem. About $42\%$ of \pws reported not using voice-based AI tools at all. Among those who do, mainstream systems (e.g., Siri) were often described as ``\emph{not at all accessible}'' or ``\emph{not very accessible}''. A recurring problem was that voice assistants stop listening during blocks, repetitions, or extended pauses. These technical failures lead to \textbf{frustration and anxiety}, causing many \pws to \emph{avoid using voice-based AI tools} in general. 

% \begin{displayquote}
%     \dq{\textit{I would change the fact that they stop listening when you start stuttering; that would be very helpful.}} -- \pws 
% \end{displayquote}

\begin{participantquote}
  \itshape\enquote{I would change the fact that they stop listening when you start stuttering; that would be very helpful.} 
  \vspace{0.5ex} \\ 
  \raggedleft \normalfont \small -- \pws
\end{participantquote}

Preferences around transcription further reveal divergent needs. When asked to rate possible speech recognition tools, \pws generally found intended speech transcription %(where repetitions and other stuttering events are omitted) 
more useful than verbatim transcription, reflecting a desire for tools that support communication and self-presentation rather than reproducing every disfluency. \slp, in contrast, rated verbatim transcription as more useful for assessment, documentation and therapy planning, where preserving disfluencies is clinically meaningful. Accordingly, most \slp gave high usefulness scores to true-to-spoken transcripts and more neutral ratings to intended transcripts.

\subsection{Beyond “if”: Preferred Analytic Tools}
Both \pws and \slp saw value in tools that simply indicate \textbf{whether} (a binary classification task) stuttering occurs in a recording. However, they consistently preferred tools that additionally indicate \textbf{when} and \textbf{where} (an event detection task) stuttering occurs. Across responses, the “when/where” tool received slightly higher usefulness ratings than the “whether-only” tool for both groups; among respondents expressing a clear utility importance, roughly $65\%$ of \pws and $80\%$ of \slp favoured the more detailed detection option.
More broadly, stakeholders consistently framed their needs in terms of tracking patterns (how stuttering varies across situations, interlocutors and time) rather than simply obtaining a single binary label. Many \pws~expressed a need for tools that can identify when stuttering increases, highlight frequently problematic words or moments, and provide practical guidance to manage stress during speaking situations. \pws~also expressed enthusiasm for fluency-support and self-monitoring tools:

\begin{participantquote}
  \itshape\enquote{A dynamic app with exercises that help regulate stuttering, that shows the moments or words where I stumble the most and helps me regulate them.} 
  \vspace{0.5ex} \\ 
  \raggedleft \normalfont \small -- \pws
\end{participantquote}

% \begin{displayquote}
%     \dq{\textit{A dynamic app with exercises that help regulate stuttering, that shows the moments or words where I stumble the most and helps me regulate them.}} -- \pws 
% \end{displayquote}

\noindent\slp framed their needs beyond speech processing tools, highlighting generative tools that can assist in creating diverse and tailored materials for intervention:

\begin{slpquote}
  \itshape\enquote{AI tool for good speech analysis} 
  \vspace{0.5ex} \\ 
  \raggedleft \normalfont \small -- \slp
\end{slpquote}

\begin{slpquote}
  \itshape\enquote{I need assistance estimating the approximate (stuttering) duration. I would like to generate concrete activities, tailored to each individual, more quickly.} 
  \vspace{0.5ex} \\ 
  \raggedleft \normalfont \small -- \slp
\end{slpquote}
    
% \begin{displayquote}
%     \dq{\textit{AI tool for good speech analysis}} -- \slp \\
%     \dq{\textit{I need assistance estimating the approximate (stuttering) duration. I would like to generate concrete activities, tailored to each individual, more quickly.}} -- \slp
% \end{displayquote}

\subsection{AI Adoption and Concerns}

Only $20\%$ of \slp feel their field is \emph{ready} for AI integration. In contrast, a majority ($63\%$) indicated their openness to adopting AI tools. A majority ($63\%$) of \slp expressed strong concerns over model accuracy, ethical factors (data privacy) and lack of technical knowledge about these tools. Interestingly, they also expressed a strong preference for explainable models. When asked how important it is to understand how a model works and why it flags particular segments as stuttering, responses were overwhelmingly at the high end of the scale (range 3–5 only; no 1s or 2s). Taken together, these responses indicate a readiness–willingness gap: \slp are open to AI, but adoption is conditional on trustworthy accuracy, privacy safeguards, and workflow fit. This makes explainability a core requirement rather than an add-on, favouring transparent, segment-level outputs (what was flagged and why) supported by practical guidance for clinicians.

\begin{slpquote}
  \itshape\enquote{Respect for and privacy of patient data. That the tools be critical and have up-to-date information on stuttering. That they analyse not only disfluencies, but also moments of avoidance and struggle.} 
  \vspace{0.5ex} \\ 
  \raggedleft \normalfont \small -- \slp
\end{slpquote}

% \begin{displayquote}
%     \dq{\textit{Respect for and privacy of patient data. That the tools be critical and have up-to-date information on stuttering. That they analyze not only disfluencies, but also moments of avoidance and struggle.}} -- \slp
% \end{displayquote}

\subsection{Expectations and Perceived Benefits From AI}
 \slp viewed collaboration with AI researchers as essential. Ratings for the importance of interdisciplinary collaboration were very high (mean around 4.5 on a 5-point scale, with many ($\sim67\%$) respondents selecting the maximum. Importantly, \slp felt collaboration should begin at the earliest stages (problem identification and project planning), rather than only at the point of evaluation or deployment. Overall, \slp most often framed AI as a way to increase time efficiency, while also supporting more objective analysis, better documentation, and improved research data collection, with additional value in innovative therapy materials and expanded access to care for \pws. 
% \begin{displayquote}
%     \dq{\textit{I would like AI tools to help organise, analyse, and structure clinical data more efficiently, so that I can focus more on direct work with clients and on tailoring interventions to their specific needs.}} -- \slp
% \end{displayquote}

\begin{slpquote}
  \itshape\enquote{I would like AI tools to help organise, analyse, and structure clinical data more efficiently, so that I can focus more on direct work with clients and on tailoring interventions to their specific needs.} 
  \vspace{0.5ex} \\ 
  \raggedleft \normalfont \small -- \slp
\end{slpquote}

% \begin{table}[h]
%  \caption{Benefits, count of SLPs who want them from AI tools and the frequency(in perecentage) of their selection by SLPs.}
%     \centering
%     \resizebox{0.8\columnwidth}{!}{%
%     \begin{tabular}{@{}lcc} \toprule
%          \textbf{Benefit} & \textbf{Count} & \textbf{Percent}\\ \midrule
%          Time efficiency	 & 22 & \simpct{73}\\
%         Access to innovative materials & 15 & $50\%$\\
%        Easier data collection for research  & 12 & $40\%$\\
%          Objectivity in analysis & 11 & \simpct{37}\\
%          Improved documentation & 7 & \simpct{23}\\
%          	More access for PWS to treatment & 6 & $20\%$\\
%          Enhanced patient engagement & 4 & \simpct{13}\\ \bottomrule
%     \end{tabular}
%      }
%     \label{tab:clinical benefit}
% \end{table}
\begin{table*}[t]
    \centering
    \small
    \setlength{\tabcolsep}{6pt}
    \caption{Stuttered-speech Research × Stakeholder Alignment}
    \vspace{-2mm}
    \renewcommand{\arraystretch}{1.2}
    \resizebox{2\columnwidth}{!}{%
    \begin{tabular}{p{0.18\textwidth} p{0.28\textwidth} p{0.28\textwidth} p{0.26\textwidth}}
    \toprule
         \textbf{Research Area} & \textbf{What most papers do} & \textbf{What stakeholders say they need }& \textbf{Alignment gap} \\ \midrule

         % \multicolumn{2}{@{}l}{\textit{Research Area - Sub-task}}&& \\ \hdashline

         % \textbf{Stutter Identification – Classification }(binary / multi-class / multi-label) & Dominant task occuring in $133$ of $228$ papers, mostly clip-level classification; often labelled “detection” even when no temporal localisation is done.  & \pws and \slp see \emph{some} value in knowing whether stuttering occurs & \textbf{Over-served} and \textbf{mis-framed}: huge focus on “whether” stuttering occurs, but little attention to when, where, and in which contexts it appears.\\ \midrule
                  \rowcolor{gray!18}
         \textbf{Stutter Identification – Detection} & Rare (only a handful of true detection papers, despite $72$ using “detection” in the title). & Both stakeholders rate “when/where” tools more useful than “whether-only”; they want patterns over time, situations, and interlocutors. & \textbf{Underserved:} stakeholders want temporal, pattern-aware tools; literature mostly offers overall static labels. \\ \midrule
         
         \textbf{Stutter Identification – Severity Assessment} & Research is dominated by classification work; A small minority of works on clinically grounded severity assessment. & \slp~ strongly want tools for objective severity tracking. \pws~ want to measure severity in real life scenarios. & \textbf{Underserved:} practically unexplored in research; identified need by both stakeholders \\ \midrule

                    \rowcolor{gray!18}
         \textbf{Speech Recognition – Verbatim ASR} & Most ASR papers use intended reference for evaluation; most ASR models are sometimes default verbatim predictors; repetitions/fillers sometimes cleaned without discussion. & \slp~ strongly prefer verbatim transcripts for  assessment, documentation, severity tracking, and clinical reasoning. & \textbf{Partially aligned but undefined:} models often behave “as if” verbatim, but papers rarely state this explicitly or discuss clinical use. \\ \midrule

         \textbf{Speech Recognition – Intended ASR} & Many systems implicitly move toward intended ASR (e.g., filler removal) without saying so. & \pws~prefer intended transcripts for everyday communication; they want systems that “listen patiently” and ignore disfluencies. & \textbf{Underserved:} intended ASR is a key \pws~ need, but mostly implicit in modelling and evaluation; not often framed as a distinct goal.\\ \midrule

                  \rowcolor{gray!18}
         \textbf{Data-Centric} (corpus curation, annotation schemes, synthetic data) & Growing area, with recent emphasis on large synthetic corpora; diverse annotation schemes and label sets. & \pws~ report high willingness to donate real speech; \slp~ stress the need for diverse, well-annotated, clinically meaningful data. & \textbf{Synthetic-heavy, no standards:} researchers lean on synthetic data while real-speech donors are available; annotation schemes often lack clinical grounding% and expertise. 
         \\ \midrule
         \textbf{Research-Focused} (evaluation, surveys, explainability, ethics, deployment) & Relatively small: a few surveys, evaluation/analysis papers, and very few on ethics, privacy, or explainability. & \slp~ explicitly ask for explainable models, ethical safeguards, and guidance on how to use outputs in practice. & \textbf{Underserved:} critical reflection, explainability, and deployment guidance lag far behind model-building. \\ \bottomrule
        %  \multicolumn{2}{@{}l}{\textit{Research Practice}}&& \\  \hdashline
        %  \textbf{Multilingual / Cross-lingual Modeling} & Strong English bias; most studies are monolingual; limited cross-lingual or low-resource work. & Both \pws and \slp operate in many languages and report language-dependent variability. & \textbf{Representation gap}: research concentrates on English, while stakeholder communities are linguistically diverse and stutter can vary by language. .\\ \midrule
        %  \textbf{High-resource Dependent Frameworks} & Increase in models built using high-resource dependent frameworks (large synthetic data, large scale pre-trained models). & \pws report stutter variability by language, context and collocutors. & \textbf{Adaptation gap}: Existing system cannot be adapted to low-resource settings due to high-resource dependency \\ \midrule
        % \textbf{Stakeholder Involvement} & $\prec20\%$ of papers report any stakeholder involvement; collaboration is usually limited to data collection. & \slp rate early, deep collaboration as very important; \pws express willingness to contribute data and feedback. & \textbf{Process gap}: models are rarely co-designed with stakeholders, despite clear appetite for involvement.\\ \midrule   
        % \textbf{Open Science} & Only $10\%$ of papers report release of code, models or data. & \pws report lack of knowledge of existence about tools. & \textbf{Adaptation gap}\\ 
    \end{tabular}
    }
    \label{tab:taxonomyAlignment}
    \vspace{-0.5em}
\end{table*}

\section{Alignment Analysis}%Results: 
\label{sec:alignment}To move beyond a stand-alone literature survey and a stand-alone stakeholder survey, we explicitly analyse alignment between research priorities and stakeholder needs in this section and interpret what these gaps imply for stuttering and research community. 

% We identify, for each research area (defined in Section~\ref{sec:ResearchAreas}), where literature priorities and stakeholder needs converge or diverge
% For each major research area and sub-task type in our taxonomy (defined in Section~\ref{sec:ResearchAreas}), we (i) quantify research area attention using paper count from our 228-paper corpus (Section \ref{sec:literatureCount}), and (ii) quantify stakeholder priorities/needs using ratings from \pws and \slp on the usefulness of corresponding tool types and the frequency of related themes in their free-text responses (Section~\ref{sec:stakeholder needs}).
%In this section we present findings from the alignment analysis in terms of alignment gaps and implications of findings on the stuttering and research community. 
% \red{to edit}

% Bringing together the literature mapping and stakeholder insights using the research taxonomy and our proposed sub-tasks reveals several specific alignment gaps between what stuttered-speech AI research currently optimises for and what \pws and \slp say they need.

%\subsection{The Alignment Gap}
\subsection{The Gaps in Alignment}
Table \ref{tab:taxonomyAlignment} summarises the main \ztedit{gaps} identified in our analysis.\\

\noindent \textbf{Stutter Identification Gaps}. 
Our taxonomy distinguishes between classification (i.e., assigning labels to segments or clips) and detection (i.e., locating events by time-stamp). 
While stakeholder survey results show a clear preference for stutter detection (``when/where'') tools over stutter classification (``whether-only'') tools, the majority of the $96$ papers that focus exclusively on stutter identification formulate the task as classification on pre-segmented audio.
% \zc{This is somewhat surprising, did they say they wanted classification on presegmented audio? Or what did they say exactly. I'm just surprised that they'd know the systems well enough.\response{H}we asked them if they prefer classification or detection with clear definition} 
Very little work (about 8\%) addresses the fine-grained detection task that stakeholders care about.
% \zc{What is the evidence that you're drawing this from? State that at the beginning of the sentence (because it's a very nicely written and sharp sentence).} 
\ztedit{Furthermore, n}\ztdelete{N}on-standardized terminology \ztdelete{additionally amplifies the confusion}\ztedit{creates additional barriers}, and paper titles and abstracts rarely make the underlying task formulation explicit. 
For example, we found that \textbf{72 papers} include \textit{\dq{detection}} in their title, but \textbf{only} \textbf{6} actually perform temporal detection under our definition.
% \zc{What do they do instead?} 
This \emph{naming mismatch} makes it harder to build shared benchmarks and for users to identify systems that match their needs.

\noindent \textbf{Speech Recognition Gaps.} Our taxonomy differentiates \textit{intended} and \textit{verbatim} ASR, and the stakeholder surveys indicate that these two forms serve distinct use-cases: \pws prefer intended transcripts for communication, while \slp rely on verbatim transcripts for clinical work. Yet very few ASR papers for atypical speech clearly state which transcription objective they optimise, or which reference transcript is treated as ground truth. 
This ambiguity arises because \emph{real} stuttered-speech resources often provide only a single reference transcript (typically intended), especially for reading scenarios, whereas synthetic sources often include both verbatim and intended references. Some \ztdelete{works}\ztedit{work} also \ztedit{performs}\ztdelete{perform} data preparation that implicitly \ztdelete{involve}\ztedit{involves the} removal of fillers or repetitions.
As a result, it is often unclear which stakeholder transcription need a system is optimising for, and the results become difficult to compare or reproduce results across datasets and studies. These implicit choices can optimize models into one objective, limiting their applicability and sometimes erasing \emph{clinically relevant} behaviour.  
% \zc{so what? Maybe put this sentence at the end of the paragraph} 

\noindent \textbf{Research-Focused Gaps.} Within the research-focused portion of the literature, we found only a handful of papers on explainable methods, ethics or deployment, compared to a larger body of work on building stutter identification or ASR models. 
Most systems report only global metrics, with little support for clinicians to inspect or interrogate individual predictions. 
From the \slp perspective, explainability is non-negotiable. 
Understanding how reliable the model is in specific contexts, and how outputs should inform clinical reasoning is essential and considered non-trivial.

\noindent \textbf{Data-Centric Gaps.} 
Recent work \cite{zhang25u_interspeech, Zhou2024YOLOStutterER} has invested heavily in large-scale synthetic stuttering corpora, while comparatively little effort goes into curating real stuttered speech. 
Yet our \pws survey shows that \pws are highly willing to donate recordings, both anonymously and otherwise to support more inclusive speech technology. This creates a striking misalignment: researchers are increasingly simulating stuttering instead of partnering with a community that is ready to share real speech, even though synthetic data cannot fully capture the intensity, variability, and contextual nuance of lived stuttering.

\section{Moving Forward}
\label{sec:conclusion}
% several fundamental challenges for stuttered-speech research
% the misalignments highlighted in Table~\ref{tab:taxonomyAlignment} are compounded by the fact that
Our analysis reveals that current research priorities and task formulations only partially overlap with what \pws and \slp identify as valuable, and fewer than $20\%$ of works report any stakeholder involvement. As a result, many research outputs remain difficult to adapt and insufficiently user-centred, limiting their usefulness in real clinical, everyday settings. Closing the gap between stuttered-speech research and stakeholders requires coordinated changes in research focus, evaluation practices and interdisciplinary collaboration. We outline concrete steps for the research community in this section.

% The misalignments highlighted in Table~\ref{tab:taxonomyAlignment} could be reduced through deeper stakeholder involvement in task formulation, annotation design, and evaluation, yet currently less than $20\%$ of works in our corpus involved stakeholders. 

% \red{Is there a better way to present this?}
\subsection{Research Directions}
\begin{enumerate}
    \item \textbf{Diversify and Standardise}: Future work should move beyond classification tasks to situation-aware modelling of stuttering, including detection tasks, models that capture variability across situations, interlocutors and languages, and tools for longitudinal severity tracking. 
    Researchers should also adopt standardised terminology for tasks and annotations, such as our proposed taxonomy, to improve research communication, benchmarks, and real-world relevance. 
    \item \textbf{\ztedit{Explicate}\ztdelete{Explicit intended vs. verbatim} Objectives}: ASR research on disfluent or stuttered speech should clearly state whether it targets intended or verbatim transcription, and design systems capable of producing both forms. 
    Dataset creators can help by documenting transcription conventions and providing parallel intended and verbatim annotations where feasible, and clearly specify the intended use cases of such annotations.
    % Underserved research-focused areas:
    \item \textbf{Underserved Research-Focused Areas}: Explainability, privacy and generalizability are currently under-researched areas despite being important to stakeholders; we urge researchers to weigh these aspects more heavily in their methodology, evaluation, and discussion.
    \item \textbf{Multilingual, Inclusive Modelling}: The field should treat multilingual support as a core requirement, not an afterthought. 
    Promising avenues include cross-lingual transfer, language-agnostic representations, and models trained on diverse datasets that reflect the languages, accents, and dialectal variations in real-life to truly \emph{speak together}.
    \item \textbf{Open-Science}: Researchers should try to publish open-source data, code, and artefacts to allow benchmarking, verification, and adaptation. This ensures that research findings can be replicated and interpreted accurately while enabling faster iterations that can accelerate research and development of real-world applications.
\end{enumerate}

\subsection{Evaluation Practices}
\noindent \textbf{Task-aware Benchmarks.} Benchmarks and shared tasks should be structured around clear distinctions. These include: \emph{Classification} vs. \emph{Detection}, \emph{Intended} vs. \emph{Verbatim} recognition, etc. This will make it easier to compare systems, match them with stakeholder needs, and perform \ztedit{appropriate} evaluations.
% \begin{itemize}
%     \item \emph{Classification} vs. \emph{Detection}
%     \item \emph{Intended} vs. \emph{Verbatim} recognition
%     \item \emph{Multi-class} vs. \emph{Multi-label} annotations
% \end{itemize} 

\noindent \textbf{Explainability and Evaluation.} For stutter identification tools, evaluation should consider how interpretable outputs are to end users. 
This can involve user studies of visualisations, segment-level explanations, and their impact on trust and decision-making. Evaluations should go beyond overall accuracy or F1-scores to include performance across different contexts (situations, interlocutors, languages). 
For ASR, evaluation should look beyond word error rates and analyse how prediction errors affect intelligibility and communication. 

% \noindent \textbf{Benchmarks and Standardization:} Another important resource are comprehensive evaluation benchmarks to test models for various tasks in a unified manner, this will improve adaptation by end-users. Also having standardized task definitions will allow correct evaluation and comparison of methods. A huge concern for SLPs is accuracy and a benchmark will allow them to evaluate available methods.

\subsection{Interdisciplinary Collaboration}
% Based on the finding that 67\% of SLPs view clinical collaboration as essential, we propose a \dq{Stakeholder-in-the-Loop} development cycle that begins at the problem-definition stage.\zc{I hate the framing of stakeholder in the loop, can we-can we call it co-design or co-research? I've proposed an alternative phrasing below. feel free to use/edit change as you see fit.}
 \ztedit{Based on the finding that 67\% of \slp view clinical collaboration as essential, research should actively seek to engage end users in their work throughout the work lifecycle. 
 We argue that collaboration will be particularly beneficial for all parties if stakeholders are included in:}
 
 \begin{itemize}
     \item \textit{Problem definition}: involve \pws~and/or \slp~from the onset to identify meaningful problem formulation and realistic use-cases.
     \item \textit{Data collection and annotation:} co-design recruitment, consent and labelling schemes with \pws~and/or \slp~ to reflect clinically relevant categories, recognize user intent, and proactively address privacy and ethics concerns.
     \item \textit{Model development:} consult stakeholders on trade-offs (for example, sensitivity vs. false alarms) and on which errors are most problematic \ztedit{for a given scenario}.
     \item \textit{Evaluation and deployment:} jointly interpret results, co-design interfaces, and plan integration into clinical workflows and everyday communication practices.
 \end{itemize}

 To support this, future work should explicitly report stakeholder roles and the stages at which they were involved, just as they report dataset size and model architecture. 
 Funders and institutions can further encourage long-term partnerships between AI researchers and end users, recognising that trustworthy, human-centred systems for atypical speech require time, mutual learning and shared ownership.

% \newpage

% Interspeech Requirement
\section{Generative AI tools}
Generative artificial intelligence tools were used solely to assist with language editing for clarity of presentation. All research questions, annotations, and interpretations were conceived and carried out by the authors, who take full responsibility for the originality, validity, and integrity of the work.

\bibliographystyle{IEEEtran}
\bibliography{mybib}

\end{document}